\documentclass{article}

\usepackage{microtype}
\usepackage{graphicx}
\usepackage{subcaption}
\usepackage{booktabs}
\usepackage{tikz}
\usetikzlibrary{positioning, shapes.geometric, arrows.meta, calc}
\usepackage{float}

\usepackage{hyperref}

\usepackage[accepted]{icml2026}
\makeatletter
\renewcommand{\Notice@String}{}
\makeatother

\usepackage{amsmath}
\usepackage{amssymb}
\usepackage{mathtools}
\usepackage{amsthm}

\usepackage[capitalize,noabbrev]{cleveref}

\theoremstyle{plain}

\theoremstyle{definition}

\theoremstyle{remark}

\icmltitlerunning{TimeRouter: Efficient and Adaptive Routing of Time-Series Foundation Models}

\begin{document}

\twocolumn[
\icmltitle{TimeRouter: Efficient and Adaptive Routing of \\
           Time-Series Foundation Models}

% \icmlsetsymbol{equal}{*}

\begin{icmlauthorlist}
\icmlauthor{Kanghui Ning}{uconn}
\icmlauthor{Yushan Jiang}{uconn}
\icmlauthor{Kashif Rasul}{ms}
\icmlauthor{Anderson Schneider}{ms}
\icmlauthor{Yuriy Nevmyvaka}{ms}
\icmlauthor{Dongjin Song}{uconn}
\end{icmlauthorlist}

\icmlaffiliation{uconn}{School of Computing, University of Connecticut, Storrs, USA}
\icmlaffiliation{ms}{Department of Machine Learning Research, Morgan Stanley, New York, USA}

\icmlcorrespondingauthor{Dongjin Song}{dongjin.song@uconn.edu}

\icmlkeywords{time-series foundation models, routing, agentic forecasting, GIFT-EVAL}

\vskip 0.3in
]

\printAffiliationsAndNotice{}
\begin{abstract}
Time-series foundation models (TSFMs) are increasingly explored as predictive experts within emerging agentic time-series systems. However, TSFMs exhibit heterogeneous inductive biases, and no single model consistently dominates across forecasting regimes, making expert selection a critical challenge. Existing systems often delegate this decision to LLM-based controllers, incurring substantial inference overhead. We present TimeRouter, an efficient routing framework that leverages empirical complementarity across a pool of pretrained TSFMs through lightweight discriminative routing, selective gating, and ensemble fallback. Concretely, TimeRouter combines a learned routing head, a selective gate, and an ensemble fallback, enabling adaptive expert selection without invoking an LLM at inference time. TimeRouter achieves state-of-the-art performance on the GIFT-EVAL leaderboard, with an LB MASE of \textbf{0.6765}. Beyond benchmark performance, our ablation studies provide empirical insights into TSFM routing design, highlighting the importance of pool composition and selective gating. Taken together, these results position TimeRouter as a modular and lightweight routing layer for future agentic time-series systems built upon foundation-model pools. Our code is available at \url{https://github.com/UConn-DSIS/TimeRouter}.
\end{abstract}

%====================================================================
\section{Introduction}
\label{sec:intro}

The last three years have seen rapid growth in time-series foundation models (TSFMs): Lag-Llama~\citep{rasul2024lagllama}, Chronos~\citep{ansari2024chronos, ansari2025chronos2}, TimesFM~\citep{das2024timesfm}, TiRex~\citep{auer2025tirex}, Moirai~\citep{woo2024moirai, liu2024moiraimoe, liu2025moirai2}, Sundial~\citep{liu2025sundial}, PatchTST-FM~\citep{wen2026patchtstfm}, FlowState~\citep{graf2025flowstate}, TTM~\citep{ekambaram2024ttm}, and others. Each is pretrained on a distinct corpus and embodies distinct architectural choices; recent analysis by~\citet{yu2025biases} identifies three design axes (patch size, embedding type, training loss) along which these choices induce orthogonal inductive biases. As a result, no single design point is optimal across the full range of forecasting regimes: which TSFM performs best varies systematically with sampling frequency, forecast horizon, domain, and noise structure. The practical challenge is therefore not identifying a universally best TSFM, but adaptively selecting the right TSFM for each input, which naturally leads to a routing problem.

\textbf{TSFMs as components in agentic systems.} TSFMs are increasingly explored as components within emerging agentic time-series systems that select or combine multiple TSFMs at inference time. TimeCopilot~\citep{garza2025timecopilot} orchestrates feature analysis and model selection via a generic LLM agent; MoiraiAgent~\citep{moiraiagent} uses a fine-tuned Qwen-2.5-3B for per-series expert selection; TSOrchestra~\citep{tsorchestra} uses an R1-style fine-tuned LLM for ensemble orchestration over a multi-FM pool. Recent non-LLM approaches explore complementary directions: Synapse~\citep{synapse} performs timestamp-level adaptive arbitration by dynamically reweighting the full TSFM pool, while ZooCast~\citep{zoocast} performs task--model matching through embedding and similarity-based Top-$K$ selection. These systems demonstrate the practical value of adaptive coordination across TSFMs, spanning LLM-based orchestration, adaptive ensemble arbitration, and embedding-based model matching. However, a lightweight discriminative routing layer for adaptive expert selection across TSFMs remains largely unexplored.

\textbf{Theoretical basis.} TimeRouter is motivated by two complementary ideas: stacked generalisation and selective prediction. \emph{Stacked generalisation}~\citep{wolpert1992stacking} trains a second-level model over first-level outputs treated as features, motivating our use of per-FM cross-validation scores and downsampled forecasts as routing features, together with a CV-inverse-weighted ensemble fallback. \emph{Selective prediction and learning-to-defer}~\citep{geifman2017selective, mozannar2020consistent, verma2022calibrated} establish that confidence-thresholded classifiers admit controllable risk--coverage tradeoffs. TimeRouter adopts this principle through a selective gate that routes low-confidence inputs to the ensemble fallback rather than committing to a single expert.

\textbf{TimeRouter.} TimeRouter uses a one-vs-all classifier head to produce routing scores over a pool of TSFMs from a feature representation combining context information and base-model outputs. At inference, the classifier's decision-space margin and the pool's forecast-space diversity jointly drive a selective gate: low-confidence inputs are deferred to a CV-inverse-weighted ensemble fallback, while confident inputs commit to the classifier's argmax. On GIFT-EVAL~\citep{gifteval2024}, TimeRouter achieves \textbf{LB MASE 0.6765}, a new state-of-the-art on the leaderboard.

\textbf{Contributions.}
(i)~We propose TimeRouter, an efficient discriminative routing framework for adaptive expert selection across time-series foundation models.
(ii)~On GIFT-EVAL, TimeRouter achieves LB MASE $0.6765$, a new state-of-the-art on the leaderboard. The ablation studies further provide empirical insights into TSFM routing design.
(iii)~TimeRouter provides a modular and efficient routing layer for future agentic time-series systems built upon foundation-model pools.

%====================================================================
\section{Method}
\label{sec:method}

\begin{figure}[t]
\centering
\includegraphics[width=0.85\columnwidth]{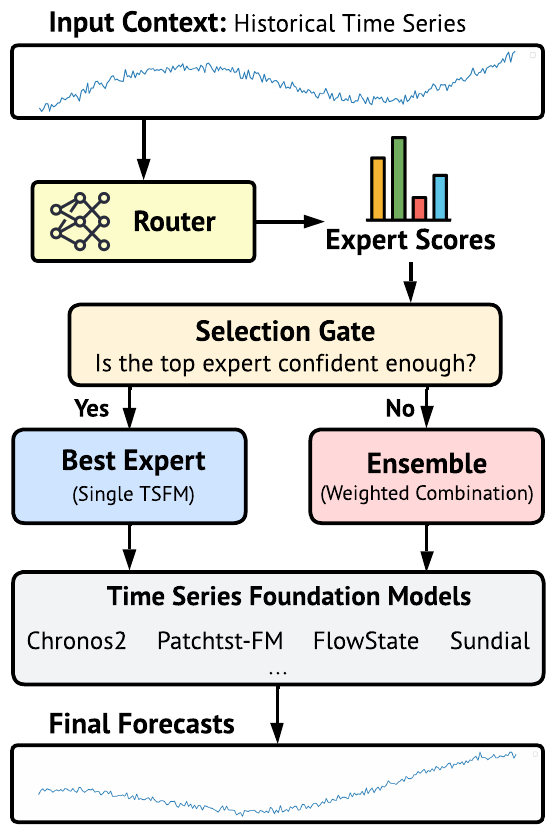}
\caption{Overview of TimeRouter. Given an input context, the router produces routing scores over a pool of time-series foundation models (TSFMs). A selective gate determines whether to commit to the top expert or defer to an ensemble fallback when confidence is low, enabling adaptive expert selection across heterogeneous forecasting regimes.}
\label{fig:arch}
\end{figure}

\textbf{Problem definition.} Given a fixed pool of $K$ frozen TSFMs $\mathcal{F} = \{F_1, \dots, F_K\}$, each $F_k$ maps a univariate context $x \in \mathbb{R}^{T}$ to a point forecast $F_k(x) \in \mathbb{R}^{H}$ over the next $H$ steps. A deterministic ensemble combiner $\mathrm{Ens}: \mathbb{R}^{H \times K} \to \mathbb{R}^{H}$ is fixed in advance. A \emph{router} is a policy $\pi(x; \mathcal{F}) \in \{1, \dots, K, \mathrm{Ens}\}$ producing the forecast
\begin{equation}
\hat{y}_\pi(x) =
\begin{cases}
F_{\pi(x)}(x) & \pi(x) \in [K], \\
\mathrm{Ens}\bigl(F_1(x), \dots, F_K(x)\bigr) & \pi(x) = \mathrm{Ens}.
\end{cases}
\label{eq:router}
\end{equation}
We seek a router that minimises the expected per-row loss
\begin{equation}
\pi^\star = \arg\min_\pi\, \mathbb{E}_{(x, y) \sim \mathcal{D}}\!\bigl[\ell\bigl(\hat{y}_\pi(x), y\bigr)\bigr],
\end{equation}
where $\mathcal{D}$ is the training distribution and $\ell$ is the per-row MASE.

\textbf{Routing head and training objective.} TimeRouter parameterises $\pi$ by a one-vs-all (OvA) classifier over a fixed feature map $\phi(x) \in \mathbb{R}^{d}$ that concatenates context statistics (trend, seasonality, autocorrelation, length, $\dots$), per-FM context-tail cross-validation scores, and downsampled per-FM forecasts; the latter two blocks are stacking-style features in the sense of~\citet{wolpert1992stacking}. For each $k \in [K]$, a binary classifier $g_{\theta_k}: \mathbb{R}^{d} \to [0, 1]$ predicts whether $F_k$ is the oracle-best FM on input $x$; denoting the oracle label
\begin{equation}
k^\star(x, y) = \arg\min_{k \in [K]} \ell\bigl(F_k(x), y\bigr),
\end{equation}
each binary classifier is trained by minimising the expected binary cross-entropy
\begin{equation}
\theta_k^\star = \arg\min_{\theta_k}\, \mathbb{E}_{(x,y)\sim\mathcal D}\bigl[\mathrm{BCE}(g_{\theta_k}(\phi(x)),\, \mathbb{1}\{k^\star = k\})\bigr].
\label{eq:bce}
\end{equation}
At inference, the $K$ classifier scores are $L_1$-normalised into a score vector $p(x) = (p_1(x), \dots, p_K(x))$. The router commits to $\arg\max_k p_k(x)$ when the gate trusts the prediction and defers to $\mathrm{Ens}$ otherwise.

Two scalar signals are computed at inference time from the routing scores and pool forecasts:
\begin{align}
\text{margin: } & m(x) = p_{(1)}(x) - p_{(2)}(x), \\
\text{diversity: } & d(x) = H^{-1} \sum_{t=1}^{H} \mathrm{std}_{k}\!\left( F_k(x; t) / s(x) \right),
\end{align}
where $p_{(j)}$ is the $j$-th order statistic of the score vector $p(x)$, $F_k(x; t)$ is FM $k$'s point forecast at horizon step $t$, and $s(x)$ is the per-series context scale. The margin lives in \emph{decision space}; diversity lives in \emph{forecast space} and is large on inputs where the FMs disagree about the future. Low diversity indicates that the pool forecasts are already highly consistent, in which case committing to a single FM offers limited advantage over the ensemble fallback. Given thresholds $(\tau_m, \tau_d) \ge 0$, the gate routes
\begin{equation}
\label{eq:gate}
\pi(x) = \begin{cases}
\mathrm{Ens} & \text{if } m(x) < \tau_m \text{ or } d(x) < \tau_d, \\
\arg\max_{k} p_k(x) & \text{otherwise}.
\end{cases}
\end{equation}
Thresholds $(\tau_m, \tau_d)$ are selected on training-split OOF (\S\ref{sec:experiments}).

\textbf{Ensemble combiner.} The deployed combiner is a CV-inverse-weighted average:
\begin{align}
\mathrm{Ens}\bigl(F_1(x), \dots, F_K(x)\bigr) &= \sum_{k=1}^{K} w_k(x)\, F_k(x), \label{eq:ens} \\
w_k(x) &\propto \frac{1}{\mathrm{CV\_score}_k(x) + \epsilon}, \notag
\end{align}
where $\mathrm{CV\_score}_k$ is the context-tail single-window CV-MASE for FM $k$ and weights are normalised to sum to one. Alternative combiners (unweighted mean, per-step median, inverse-CRPS) plug into the same gate without changing the head or threshold-tuning procedure.

%====================================================================
% \vspace{-2mm}
\section{Experiments}
\label{sec:experiments}

\textbf{Benchmark and pool.} We evaluate on GIFT-EVAL~\citep{gifteval2024}, a $97$-task forecasting benchmark with a public leaderboard. As the foundation-model pool, we use four checkpoints from multiple forecasting paradigms: Chronos-2~\citep{ansari2025chronos2}, FlowState~\citep{graf2025flowstate}, PatchTST-FM~\citep{wen2026patchtstfm}, and Sundial~\citep{liu2025sundial}. These models are chosen for their strong standalone leaderboard performance and complementary forecasting behaviour. All pool members remain frozen during routing-head training.

\textbf{Implementation details.} The feature map has $d{=}305$ dimensions for our four-FM pool ($165$ pool-independent dimensions plus $35$ per FM, of which $3$ are CV statistics and $32$ are forecast-snippet buckets; block-level breakdown in Appendix~\ref{app:impl}). The one-vs-all classifier uses XGBoost~\citep{chen2016xgboost} for each binary $g_{\theta_k}$, one per FM. We fit $S{=}5$ seeds differing only in \texttt{random\_state}; each seed's $K$ classifier scores are $L_1$-normalised into a per-seed score vector and the $S$ per-seed vectors are averaged at inference time. The gate thresholds $(\tau_m, \tau_d)$ are selected by a grid sweep on a 5-fold task-grouped \texttt{GroupKFold} OOF split.

\textbf{Main result.} Table~\ref{tab:main} compares TimeRouter against the strongest single-FM and routing baselines on the GIFT-EVAL leaderboard. TimeRouter achieves \textbf{LB MASE $0.6765$}, a new state-of-the-art on the leaderboard; it improves over the strongest single FM (Chronos-2 at $0.6978$) by $\sim$$200$~bp and edges the strongest LLM-judge router (TSOrchestra at $0.6768$) by $\sim$$3$~bp while incurring no LLM inference cost at the foundation-model routing step.

\begin{table}[t]
\centering
%\footnotesize
\small
\setlength{\tabcolsep}{4pt}
\begin{tabular}{@{}lc@{}}
\toprule
Method & LB MASE \\
\midrule
\multicolumn{2}{@{}l}{\emph{Foundation models (single)}}\\
Chronos-2~\citep{ansari2025chronos2}    & 0.6978 \\
TimesFM-2.5~\citep{das2024timesfm}      & 0.7050 \\
PatchTST-FM~\citep{wen2026patchtstfm}   & 0.7069 \\
TiRex~\citep{auer2025tirex}              & 0.7158 \\
FlowState~\citep{graf2025flowstate}     & 0.7262 \\
Moirai2~\citep{liu2025moirai2}          & 0.7281 \\
Sundial~\citep{liu2025sundial}          & 0.7502 \\
TTM-R2~\citep{ekambaram2024ttm}          & 1.0196 \\
\midrule
\multicolumn{2}{@{}l}{\emph{Agentic systems}}\\
TSOrchestra~\citep{tsorchestra}          & 0.6768 \\
MoiraiAgent~\citep{moiraiagent}          & 0.6887 \\
Credence (leaderboard entry)             & 0.6907 \\
ZooCast~\citep{zoocast}                  & 0.6920 \\
Synapse~\citep{synapse}                  & 0.6986 \\
TimeCopilot~\citep{garza2025timecopilot} & 0.7051 \\
\midrule
\textbf{TimeRouter}              & \textbf{0.6765} \\
\bottomrule
\end{tabular}
\vspace{2mm}
\caption{GIFT-EVAL LB MASE comparison (lower is better). LB MASE is the geomean across $97$ tasks of (method MASE $/$ Seasonal-Naive MASE).}
\label{tab:main}
\vspace{-5mm}
\end{table}

\textbf{Efficiency.} 
TimeRouter trains in $\sim110$~s for a $20$-expert pool ($20$ one-vs-all XGBoost classifiers; $\sim155$K training rows) and incurs only $9.9$~ms inference overhead per series. 
Its lightweight routing design enables rapid adaptation as the TSFM ecosystem evolves: newly released foundation models can be added to the pool and the routing head re-trained in minutes, without updating any TSFM or invoking an LLM-based orchestration loop. Compared with recent agentic-routing systems, TimeRouter substantially reduces routing overhead while preserving the flexibility of multi-expert coordination. Additional implementation and hardware details are provided in Appendix~\ref{app:efficiency}.

\begin{table}[h]
\centering
\footnotesize
\setlength{\tabcolsep}{4pt}
\begin{tabular}{@{}p{2.0cm}p{3.4cm}p{1.7cm}@{}}
\toprule
Method & Additional learning cost & Routing latency / series \\
\midrule
TSOrchestra & fine-tune 3B LLM & $\geq472.6$ ms \\
MoiraiAgent & fine-tune 3B LLM         & $472.6$ ms \\
TimeCopilot & no training; LLM agent   & $\geq450$ ms \\
Synapse     & training-free & --- \\
ZooCast     & train embedding selector & --- \\
\midrule
\textbf{TimeRouter} & $\sim$$110$~s & $9.9$ ms  \\
\bottomrule
\end{tabular}
\vspace{2mm}
\caption{Efficiency comparison against published routing and agentic forecasting systems (TSOrchestra~\citep{tsorchestra}, MoiraiAgent~\citep{moiraiagent}, TimeCopilot~\citep{garza2025timecopilot}, Synapse~\citep{synapse}, ZooCast~\citep{zoocast}). Routing latency excludes foundation-model forecasting time for all methods. }
\label{tab:efficiency-main}
\end{table}

%====================================================================
\section{Ablations}
\label{sec:ablations}

\subsection{Selective gate ablation}
\label{sec:gate-ablation}

\begin{figure}[H]
\centering
\includegraphics[width=\columnwidth]{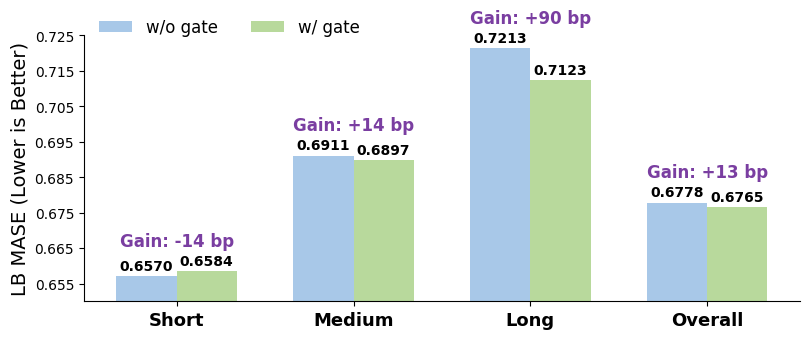}
\caption{Gate ablation on TimeRouter, stratified by GIFT-EVAL term (LB MASE, lower is better). Same head and fallback in both variants.}
\label{fig:gate-ablation}
\end{figure}

Holding the one-vs-all classifier and the CV-inverse-weighted fallback fixed, we compare the deployed selective gate against a no-gate variant that always commits to the classifier argmax. Figure~\ref{fig:gate-ablation} reports per-term LB MASE on GIFT-EVAL.

The selective gate improves overall performance by $+13$~bp, but the effect is highly term-dependent: it improves long-horizon tasks by $+90$~bp and medium-horizon tasks by $+14$~bp, while slightly regressing on short-horizon tasks ($-14$~bp). This matches the intuition that forecast uncertainty grows with horizon, making cross-FM disagreement more informative on longer-horizon inputs and ensemble fallback more beneficial there. On short horizons, the classifier argmax is already strong and the gate provides limited additional benefit.

% \begin{table}[H]
% \centering
% \footnotesize
% \setlength{\tabcolsep}{4pt}
% \begin{tabular}{@{}lcccc@{}}
% \toprule
% Variant & Short & Medium & Long & Overall \\
%         & $n{=}55$ & $n{=}21$ & $n{=}21$ & $n{=}97$ \\
% \midrule
% w/o gate                         & \textbf{0.6570} & 0.6911 & 0.7213 & 0.6778 \\
% w/ gate     & 0.6584 & \textbf{0.6897} & \textbf{0.7123} & \textbf{0.6765} \\
% \midrule
% $\Delta$ (bp)   & $-14$  & $+14$  & $+90$            & $+13$           \\
% \bottomrule
% \end{tabular}
% \caption{Gate ablation on TimeRouter, stratified by GIFT-EVAL term (LB MASE, lower is better). Same head and fallback in both rows.}
% \label{tab:gate-ablation}
% \end{table}

\vspace{-3mm}

\begin{table}[H]
\centering
\footnotesize
\setlength{\tabcolsep}{4pt}
\begin{tabular}{@{}lcc@{}}
\toprule
Head & LB MASE & $\Delta$ (bp) \\
\midrule
XGBoost(deployed)             & 0.6765 & ---  \\
LightGBM                             & \textbf{0.6762}         & $-3$ \\
Random Forest                        & 0.6776         & $+11$ \\
MLP                      & 0.6787         & $+22$ \\
Logistic Regression                  & 0.6836         & $+71$ \\
\bottomrule
\end{tabular}
\vspace{2mm}
\caption{Head ablation on TimeRouter. Same features, one-vs-all structure, and gate; only the binary classifier varies. $\Delta$ is in bp relative to XGBoost.}
\label{tab:head-ablation}
\end{table}
\vspace{-8mm}
\subsection{Routing head ablation}
\label{sec:head-ablation}

% We compare five binary classifier families for the OvA stage (Table~\ref{tab:head-ablation}), all sharing the same feature map, OvA structure, and selective gate. The four non-linear families---XGBoost, LightGBM, Random Forest, and a 2-layer MLP---cluster within a narrow 25~bp band ($0.6762$--$0.6787$), while logistic regression, the only linear classifier, degrades substantially ($+71$~bp behind XGBoost). This suggests that routing performance depends more on modeling feature interactions than on the specific non-linear classifier family. We therefore retain XGBoost as the default routing head, although the differences among the non-linear variants are practically negligible.
We compare five binary classifier families for the one-vs-all stage (Table~\ref{tab:head-ablation}), all sharing the same feature map, one-vs-all structure, and selective gate. The four non-linear families, XGBoost, LightGBM, Random Forest, and a 2-layer MLP, cluster within a narrow 25~bp band ($0.6762$--$0.6787$), while logistic regression, the only linear classifier, degrades substantially ($+71$~bp behind XGBoost). This suggests that routing performance depends primarily on modeling non-linear feature interactions rather than on the specific classifier family. We therefore use XGBoost as the default routing head.

\subsection{Pool composition ablation}
\label{sec:pool-ablation}

To assess the contribution of each pool member, we ablate the four-FM pool by evaluating all $\binom{4}{3}{=}4$ three-FM subsets and all $\binom{4}{2}{=}6$ two-FM subsets. Table~\ref{tab:pool-ablation} reports results for every subset, each using the same one-vs-all classifier, selective gate, and ensemble fallback as the full four-FM configuration.

% \begin{table}[H]
% \centering
% \footnotesize
% \setlength{\tabcolsep}{3pt}
% \begin{tabular}{@{}p{5.2cm}c@{}}
% \toprule
% Pool & LB MASE \\
% \midrule
% \textbf{Our four-FM pool (\S\ref{sec:experiments})} & \textbf{0.6765} \\
% \midrule
% \multicolumn{2}{@{}l}{\emph{$K{=}3$ subsets (drop one FM)}}\\
% drop FlowState   & 0.6807 \\
% drop PatchTST-FM & 0.6814 \\
% drop Sundial     & 0.6834 \\
% drop Chronos-2   & 0.6876 \\
% \midrule
% \multicolumn{2}{@{}l}{\emph{$K{=}2$ subsets (keep two FMs)}}\\
% Chronos-2 + Sundial          & 0.6847 \\
% Chronos-2 + PatchTST-FM      & 0.6863 \\
% Chronos-2 + FlowState        & 0.6888 \\
% FlowState + PatchTST-FM      & 0.6889 \\
% PatchTST-FM + Sundial        & 0.6938 \\
% FlowState + Sundial          & 0.7082 \\
% \bottomrule
% \end{tabular}
% \caption{Pool ablation by reducing our four-FM pool to all three- and two-FM subsets.}
% \label{tab:pool-ablation}
% \end{table}

\begin{table}[H]
\centering
\footnotesize
\setlength{\tabcolsep}{3pt}
\begin{tabular}{@{}p{5.2cm}c@{}}
\toprule
Pool & LB MASE \\
\midrule
\textbf{Full four-FM pool} & \textbf{0.6765} \\
\midrule
\multicolumn{2}{@{}l}{\emph{$K{=}3$ subsets}}\\
drop FlowState   & 0.6807 \\
drop PatchTST-FM & 0.6814 \\
drop Sundial     & 0.6834 \\
drop Chronos-2   & 0.6876 \\
\midrule
\multicolumn{2}{@{}l}{\emph{$K{=}2$ subsets}}\\
Chronos-2 + Sundial          & 0.6847 \\
Chronos-2 + PatchTST-FM      & 0.6863 \\
Chronos-2 + FlowState        & 0.6888 \\
FlowState + PatchTST-FM      & 0.6889 \\
PatchTST-FM + Sundial        & 0.6938 \\
FlowState + Sundial          & 0.7082 \\
\bottomrule
\end{tabular}
\vspace{2mm}
\caption{Pool ablation over all three- and two-FM subsets.}
\label{tab:pool-ablation}
\vspace{-4mm}
\end{table}

\textbf{Chronos-2 is the critical anchor.} Dropping Chronos-2 yields the worst $K{=}3$ subset ($0.6876$, $+111$~bp behind the full pool); dropping any of the other three FMs costs only $+42$ to $+69$~bp. At $K{=}2$ the asymmetry persists: all three subsets that retain Chronos-2 score $0.6847$--$0.6888$, while every subset without Chronos-2 scores $\geq 0.6889$, with FlowState+Sundial the worst at $0.7082$. Pool size matters too, best-to-best LB MASE degrades from $0.6765$ ($K{=}4$) to $0.6807$ (best $K{=}3$) to $0.6847$ (best $K{=}2$), each step costing $\sim$$40$~bp.

%====================================================================
\section{Conclusion}
\label{sec:conclusion}

We presented TimeRouter, an efficient discriminative routing framework for adaptive expert selection across time-series foundation models, achieving state-of-the-art LB MASE on the GIFT-EVAL leaderboard without invoking an LLM at routing time. Our ablation studies provide several insights for future TSFM routing systems: pool composition and pool size both matter; the selective gate is most beneficial on long-horizon tasks where forecast disagreement becomes informative; and classifier family matters little as long as non-linear feature interactions can be modeled effectively. Because the routing layer is lightweight and easy to adapt, newly released TSFMs can be integrated into the pool without retraining existing foundation models, positioning TimeRouter as a modular routing layer for future agentic time-series systems.

% We presented TimeRouter, an efficient discriminative router over a frozen pool of pretrained TSFMs that achieves state-of-the-art LB MASE on the GIFT-EVAL leaderboard without invoking an LLM at the foundation-model routing step. Three ablation findings inform future TSFM routing system design: both pool composition and pool size matter; the selective gate contributes most on long-horizon tasks where cross-FM forecast disagreement is informative; and the classifier choice matters little as long as it can model non-linear feature interactions. Because the routing layer is lightweight and inexpensive to adapt, newly released TSFMs can be absorbed into the pool without retraining any foundation model, positioning TimeRouter as a modular routing layer for the next generation of agentic time-series systems.

\newpage
\bibliography{example_paper}
\bibliographystyle{icml2026}

%====================================================================
\appendix
\onecolumn
\section{Implementation Details}
\label{app:impl}

\textbf{XGBoost head.} \texttt{binary:logistic} with \texttt{max\_depth=6}, \texttt{n\_estimators=200}, \texttt{learning\_rate=0.05}, \texttt{subsample=0.8}, \texttt{tree\_method=hist}; one binary classifier per FM. We fit $S{=}5$ seeds differing only in \texttt{random\_state}; each seed produces $K$ class scores that are $L_1$-normalised into a per-seed score vector, and the $S$ per-seed vectors are averaged at inference time.

\textbf{Training data and filter.} Training rows are drawn from the GIFT-EVAL training split (disjoint from the $97$-task test split). Per-FM forecasts, oracle-best labels, and the feature vector $\phi(x)$ are precomputed once and shipped as parquet files, so the router sees only $(\phi(x), k^\star(x, y))$ at training time. We drop rows with $\mathrm{best\_mase} > 1.0$ (poor router headroom), $\mathrm{scale} < 0.01$ (degenerate near-constant series), or fewer than $2$ pool members producing a valid forecast; this leaves $\sim$$155$K training rows over $\sim$$93$ tasks for the four-FM pool.

% \textbf{Gate threshold sweep.} We sweep $\tau_m \in \{0.00, 0.02, 0.05, 0.07, 0.09, 0.10, 0.12, 0.15, 0.20, 0.25\}$ and $\tau_d \in \{0.0, 0.01, 0.02, 0.03, 0.04, 0.05, 0.07, 0.10\}$, evaluating each $(\tau_m, \tau_d)$ on a 5-fold task-grouped \texttt{GroupKFold} OOF split. The selected thresholds for the four-FM pool are $(\tau_m, \tau_d) = (0.12, 0.02)$, with test gate-rate $56\%$.

\textbf{Gate threshold sweep.} We sweep $\tau_m \in \{0.00, 0.02, 0.05, 0.07, 0.09, 0.10, 0.12, 0.15, 0.20, 0.25\}$ and $\tau_d \in \{0.0, 0.01, 0.02, 0.03, 0.04, 0.05, 0.07, 0.10\}$, evaluating each $(\tau_m, \tau_d)$ on a 5-fold task-grouped \texttt{GroupKFold} OOF split over the training set only. The selected thresholds for the four-FM pool are $(\tau_m, \tau_d) = (0.12, 0.02)$ and are fixed for all test-time evaluations.

\textbf{Ensemble combiner $\mathrm{Ens}$.} CV-inverse-weighted average $\mathrm{Ens}(F_1, \dots, F_K) = \sum_k w_k\, F_k$ with $w_k \propto 1/(\mathrm{CV\_score}_k + 10^{-8})$ and $\sum_k w_k = 1$. The per-FM CV scores are computed from context-tail validation windows for both training and test inputs and used as routing features as well as ensemble weights.

\textbf{Feature map $\phi$.} The feature vector for each $(\text{series}, \text{cutoff})$ row concatenates four blocks; total dimension $d = 165 + 35K$ ($d{=}305$ for the four-FM pool). (i) \emph{Context-window statistics} ($31$ dims): $18$ time-series descriptors (mean, std, range, IQR, skewness, kurtosis, $\mathrm{acf}$ at lags $1, 5, 10$, trend slope, diff statistics, zero-crossings, turning points, log length), $5$ static metadata (horizon, series length, horizon/length ratio, frequency descriptor, number of available pool members), and $8$ regime-shift descriptors. (ii) \emph{Normalised context snippet} ($128$ dims): the input context normalised by its own mean/std and linearly resampled to $128$ buckets. (iii) \emph{Per-FM CV statistics} ($3K + 6$ dims): for each pool member, a single-window context-tail CV-MASE score, its rank within the pool, and its gap to the pool best ($3K$ dims), plus $6$ pool-level CV aggregates. (iv) \emph{Per-FM forecast snippets} ($32K$ dims): each pool member's forecast linearly resampled to $\mathrm{PRED\_LEN}=32$ buckets and normalised by the context mean/std. Blocks (iii) and (iv) are stacking-style features in the sense of \citet{wolpert1992stacking}.

\section{Efficiency Comparison}
\label{app:efficiency}

Table~\ref{tab:efficiency} compares additional learning cost, router scale, inference latency, and LB MASE against published routing and agentic forecasting systems. Compared with LLM-based orchestration approaches, TimeRouter replaces heavyweight reasoning agents with a lightweight discriminative router that trains in minutes on CPU hardware and serves at millisecond-level latency, while achieving state-of-the-art LB MASE among publicly reproducible methods on GIFT-EVAL.

\begin{table}[h]
\centering
\small
\setlength{\tabcolsep}{4pt}
\begin{tabular}{@{}p{2.8cm}p{3.8cm}p{1.7cm}p{1.5cm}c@{}}
\toprule
Method & Additional learning cost & Router size & Infer. & LB MASE \\
\midrule
TSOrchestra~\citep{tsorchestra}
& fine-tune reasoning LLM (R1-style)
& 3B LLM
& $\geq 472.6$ ms
& 0.6768 \\

MoiraiAgent~\citep{moiraiagent}
& fine-tune 3B LLM
& 3B LLM
& 472.6 ms
& 0.6887 \\

TimeCopilot~\citep{garza2025timecopilot}
& no training; LLM agent
& LLM-scale
& $\geq 450$ ms
& 0.7051 \\

Synapse~\citep{synapse}
& training-free adaptive ensemble
& ---
& ---
& 0.6986 \\

ZooCast~\citep{zoocast}
& train co-embedding selector
& ---
& ---
& 0.6920 \\

\midrule
\textbf{TimeRouter (ours)}
& $\sim110$~s CPU
& $1.72$~MB
& \textbf{9.9 ms}
& \textbf{0.6765} \\
\bottomrule
\end{tabular}
\caption{
Additional learning cost, router size, inference latency, and LB MASE against routing-based and agentic forecasting systems. 
Dashes indicate quantities that are either not reported or not directly applicable.
}
\label{tab:efficiency}
\end{table}
Additional learning cost descriptors are sourced from each baseline's original paper or blog and are not directly comparable across methods. Inference costs are measured from publicly available implementations when reproducible code is available; otherwise, they are estimated from the reported system configuration. Our measurements are conducted on a single Intel Xeon Gold 5318Y CPU with 16 threads per classifier using XGBoost 3.1.3.

\subsection{Routing-overhead Benchmark Details}
\label{app:inference-details}

Table~\ref{tab:efficiency} compares routing/orchestration overhead only and excludes foundation-model forecasting time for all methods. Since different systems invoke different TSFM pools and forecasting backbones, including foundation-model forward passes would confound the comparison with unrelated implementation differences. We therefore isolate the latency introduced by the routing/orchestration layer itself.

All methods are evaluated on the same five GIFT-EVAL tasks: \texttt{ett1/W}, \texttt{saugeenday/M}, \texttt{us\_births/M}, \texttt{M\_DENSE/D}, and \texttt{hierarchical\_sales/W}. Following the standard rolling-window evaluation protocol, these tasks contain a total of 585 forecasting rows. We report the rows-weighted average routing latency across all five tasks.

\paragraph{TimeRouter (ours).}
We measure routing overhead only, including feature extraction and router inference, excluding all foundation-model forward passes. Measurements are conducted using a CPU-based XGBoost router (XGBoost 3.1.3) on a single Intel Xeon Gold 5318Y CPU with 16 threads per classifier.

\paragraph{MoiraiAgent.}
MoiraiAgent publicly releases its inference pipeline, allowing direct reproduction. We reproduce the routing overhead on a single NVIDIA RTX A6000 GPU under the same evaluation setup. The reported latency isolates the LLM-agent orchestration component and excludes foundation-model forecasting time.

\paragraph{TSOrchestra.}
TSOrchestra does not release reproducible inference code for its LLM-routing version. The paper reports using the same Qwen2.5-3B reasoning model as MoiraiAgent. Since TSOrchestra additionally involves multi-round reasoning and orchestration, we conservatively estimate its routing overhead to exceed that of MoiraiAgent.

\paragraph{TimeCopilot.}
TimeCopilot does not release the exact GIFT-EVAL inference configuration used in the paper. The system reports using GPT-4o-mini as the orchestration LLM. Public benchmarks report a time-to-first-token latency of roughly $0.45$--$0.5$ seconds for GPT-4o-mini, excluding additional tool-execution overhead. We therefore conservatively estimate the routing overhead to exceed $450$ ms per forecasting row.

\paragraph{Synapse and ZooCast.}
Neither Synapse nor ZooCast publicly release reproducible inference code, and the implementation details provided in the papers are insufficient to reliably isolate routing overhead from forecasting execution. We therefore mark routing latency as unavailable for both methods.

\end{document}